\documentclass{article}
\usepackage{spconf,amsmath}
\usepackage{graphicx}
\usepackage{enumerate}
\usepackage{pgfplots}
\usepackage{mathtools}
\usepackage{lipsum}
\usepackage{amssymb}
\usepackage{float}
\usepackage{multicol}
\usepackage{xspace}
\usepackage{lettrine}
\usepackage{tcolorbox}
\usepackage{subfigure}
\usepackage{tikz}
\usetikzlibrary{backgrounds,automata}
\usepackage{latexsym}
\usetikzlibrary{decorations.markings}
\usepackage{fancybox}
\usepackage[]{algorithm2e}
\usepackage[english]{babel}
\usepackage{amsthm}
\newtheorem{theorem}{Theorem}

\newcounter{alg}
\newcommand{\alg}[1]{\refstepcounter{alg}\label{#1}}

 \title{TRAINING GENERATIVE ADVERSARIAL NETWORKS WITH WEIGHTS}

\name{Yannis Pantazis$^{1}$, Dipjyoti Paul$^{2}$, Michail Fasoulakis$^{3}$\thanks{The second author is supported by the EU's H2020 research and innovation  programme under the  MSCA GA 67532 and the third author is supported by the Stavros Niarchos-FORTH post-doc fellowship for project ARCHERS. Emails: pantazis@iacm.forth.gr, \{dipjyotipaul, yannis\}@csd.uoc.gr, mfasoul@ics.forth.gr}, Yannis Stylianou$^{2}$}

\address{$^{1}$Institute of Applied and Computational Mathematics, FORTH, Greece\\
$^{2}$Department of Computer Science, University of Crete, Greece\\
$^{3}$Institute of Computer Science, FORTH, Greece}

\begin{document}
\maketitle

\begin{abstract}
The impressive success of Generative Adversarial Networks (GANs)  is often overshadowed by the difficulties in their training. Despite the continuous efforts and improvements, there are still open issues regarding their convergence properties. In this paper, we propose a simple training variation where suitable weights are defined and assist the training of the Generator. We provide theoretical arguments why the proposed algorithm is better than the baseline training in the sense of speeding up the training process and of creating a stronger Generator. Performance results showed that the new algorithm is more accurate in both synthetic and image datasets resulting in improvements ranging between 5\% and 50\%.
\end{abstract}


\section{Introduction}
\label{sec:intro}
A fully data-driven paradigm in conducting science has been emerged during the last years with the advent of GANs \cite{GPMXWOCB14}. A GAN offers a new methodology for drawing samples from an unknown distribution where only samples from this distribution are available making them one of the hottest areas in machine learning/artificial intelligence research. Indicatively, GANs have been successfully utilized in (conditional) image creation \cite{mirza2014conditional,radford2015unsupervised,odena2016conditional}, generating very realistic samples \cite{karras2017progressive,brundage2018malicious}, speech signal processing \cite{pascual2017segan,saito2018statistical}, natural language processing \cite{che2017maximum} and astronomy \cite{schawinski2017generative}, to name a few.

A GAN is a two-player \emph{zero-sum game} \cite{GPMXWOCB14,OR94} between a Discriminator and a Generator, both being powerful neural networks. They are simultaneously trained to achieve a \emph{Nash equilibrium} of the game, where the Discriminator cannot distinguish the real and the fake samples while the Generator has learned the unknown distribution. It is well-known that the training procedure of GANs often fails and several specific heuristics and hacks have been devised \cite{salimans2016improved} along with general-purpose acceleration techniques such as batch normalization \cite{ioffe2015batch}. To alleviate the difficulties of training, extensions and generalizations stemming from the utilization of a different loss function has been proposed. For instance, f-GAN \cite{nowozin2016f} is a generalization where the $f$-divergence is used instead of the \emph{Shannon-Jensen divergence} of the original GAN. Another widely-applied extension is \emph{Wasserstein GAN} \cite{arjovsky2017wasserstein} which has been further improved in \cite{gulrajani2017improved}. On the other hand, there are relatively few studies that aim directly to improve the convergence speed of training of an existing GAN.

In this paper, instead of proposing a new GAN architecture or a new GAN loss function we propose a new training algorithm inspired by the \emph{multiplicative weight update method} (MWUM) \cite{AHK12}. Our goal is to improve the training of the Generator by transferring ideas from Game Theory. Intuitively, the new algorithm puts more weight to fake samples that are more probable to fool the Discriminator and simultaneously reduces the weight of samples that are confidently discriminated as fake.
Our contributions are summarized as follow: (i) By adding weights to the training of GANs, we manage to improve the training performance with minor additional computational cost. The new approach is called \emph{Weighted Generative Adversarial Network} (WeGAN). (ii) We provide rigorous arguments that the weights of WeGAN locally reduces the loss function more or at least as much as the equally-weighted stochastic gradient descent for the Generator. (iii) The proposed algorithm is not specific to vanilla GAN \cite{GPMXWOCB14}, but it is directly transferable to other extensions such as \emph{conditional GANs}, \emph{Wasserstein GAN} and \emph{f-GAN}. This is an important generalization property of WeGAN.

Before proceeding, it is worth-noting that training methods utilizing weights for the Generator have been recently proposed \cite{hjelm2017boundary,che2017maximum,hu2017unifying}. These methods are essentially equivalent since they assign importance weights to the generated samples in order to obtain a tighter lower bound for their variational formula. However, importance weights GAN (IWGAN) cannot be applied to any type of objective function and additionally they might diverge due to their unboundedness. We implemented IWGAN and present its performance in the Results section comparing it to our algorithm.

\section{Preliminaries}
\label{Preliminaries}
{\bf GAN formulation.}
Let $G$ be the Generator and $D$ be the Discriminator of the GAN \cite{GPMXWOCB14}. $D(x)$ is the probability estimate that sample $x$ is real while $G(z)$ is the sample output of the Generator giving a noise sample $z$. In order to be trained, the following objective function of the two-player \emph{zero-sum game} has to be optimized:
\begin{equation*}
\min_G\max_D \mathbb E_{x\sim p_\text{data}} [\log D(x)] + \mathbb E_{z\sim p_z} [\log (1-D(G(z)))],
\end{equation*}
where $p_\text{data}$ is the distribution to be learned, while $p_z$ is the noise input distribution. Typically, an optimum (\emph{Nash equilibrium}) of this \emph{zero-sum game}, which is a saddle point, is estimated using \emph{stochastic gradient descent}. As it was proved in \cite{GPMXWOCB14}, the global optimum of this \emph{zero-sum game} is the point where $D(x) = 1/2$ for any sample $x$ and the Generator generates samples according to the real distribution.

\noindent
{\bf MWUM basics.}
MWUM is a classic algorithmic technique with numerous applications. 
The main idea behind this method is the existence of a number of "experts" that give some kind of advice to a decision maker. To any "expert" a specific weight is assigned and the initial weights are equal for any "expert". Then, the decision maker takes the decision according to the advice of the "experts" taking into account the weight of any of them. After this the weights are multiplicatively updated according to the performance of the advice of any individual "expert", increasing the weights of the "experts" with good performance and decreasing them otherwise and so on.
We continue with the description of our algorithm and the connection to this method.
\begin{figure}[!htb]
\small
\centering    
\alg{algorithm}
\begin{tcolorbox}[title=Algorithm~\ref{algorithm}]
\label{mwgan:alg}
\For{number of iterations}{
\For{k steps}{
Sample $\{x_1, \ldots, x_m\}$ from the data distribution $p_{data}(x)$.\\
Sample $\{z_1, \ldots, z_m\}$ from the input distribution $p_{z}(z)$. \\
Update the Discriminator by ascending its stochastic gradient:
{\footnotesize \begin{equation*} 
{\nabla_{\theta_{d}}\frac{1}{m}\sum_{i=1}^{m}\left[\log D \left(x_i\right)+\log\left(1-D\left(G\left(z_i\right)\right)\right)\right].}
\end{equation*}}
}
Sample $\{z_1, \ldots, z_m\}$ from the input distribution $p_{z}(z)$.\\
Compute the unnormalized weights:\\ 
\[\footnotesize
w_i=\eta^{\left(1-D\left(G\left(z_i\right)\right)\right)}, \ i=1,...,m.
\]
Normalize:\\ 
\[\footnotesize
w_i = \frac{w_i}{\sum_{j=1}^m w_j}, \ i=1,...,m.
\] 
Update the Generator by descending its stochastic gradient:
\begin{equation*}\footnotesize
\nabla_{\theta_{g}}\sum_{i=1}^{m}w_i\log\left(1-D\left(G\left(z_i\right)\right)\right).
\end{equation*}
}
\end{tcolorbox}
\caption{ \small Stochastic gradient ascend/descent training of WeGAN. For a direct comparison with the original GAN, we follow the formulation of \cite{GPMXWOCB14}.}
\end{figure}

\section{Weighted GAN algorithm}
\label{Weighted GAN algorithm}
\label{WeGAN:sec}
The proposed algorithm presented in Fig.~\ref{mwgan:alg} is a modification of the original GAN training algorithm. Inspired by the MWUM, instead of equally-weighted 'fake' samples, we assign a weight to each sample (the "expert" in MWUM) which multiplies the respective gradient term of the Generator. 
The weighting aims to put more strength to samples that fool the Discriminator and thus are closer to the real data. Indeed, when $D(G(z))=0$ and the Discriminator understands that the sample is fake the weight decreases by a factor $\eta \in (0,1]$. On the other hand, when $D(G(z))=1$ the weight remains the same and after the normalization step it has a value greater or equal than the previous one.
Notice also that the weights in Algorithm~\ref{mwgan:alg} depend only of the current value of the Discriminator while in the standard MWUM the weights are updated cumulatively. This modification was necessary because the input samples are different at each iteration. Indeed, new samples are generated and there is no obvious map between the current samples and the samples from the previous iteration.

\subsection{Theoretical properties of WeGAN algorithm}

A key assumption of our algorithm as well as in other weighting algorithms is that the Discriminator is faithful in the sense that it produces sound decisions for both real and fake samples. Quantitatively, it means that the Discriminator should return on average values above 0.5 when the sample comes from the real distribution and below 0.5 when fake samples are fed to the Discriminator. 
Next, we show that for a fixed Discriminator, the optimal Generator with weights as in Algorithm 1 achieves lower or equal loss value than the optimal Generator with equally-weighted samples. Hence, we expect that the inferred Generator is stronger favorably affecting the speed of convergence.

\begin{theorem}
Fix Discriminator $D$ and let $G_{D;w}^*$ and $G_{D;\frac{1}{m}}^*$ be the respective optimum Generator under weighted and equally-weighted loss function defined by
\begin{equation*} \small
L(G,D;w) = \frac{1}{m} \sum_{i=1}^m \log(D(x_i)) + \sum_{i=1}^m w_i \log(1-D(G(z_i))).
\end{equation*}
Let the weight vector, $w$, be defined according to Algorithm 1 then
\begin{equation} \small
L(G_{D;w}^*,D;w) \leq L(G_{D;\frac{1}{m}}^*,D;\frac{1}{m}).
\end{equation}
\end{theorem}

\noindent
\textit{Proof.} By definition, it holds for the optimum Generator that
$$ \small
L(G_{D;w}^*,D;w) \leq L(G_{D;\frac{1}{m}}^*,D;w).
$$
If we prove that for any $G$, it holds that $L(G,D;w) \leq L(G,D;\frac{1}{m})$ when $w$ is defined as in Algorithm 1 we are done because we get the desired result for $G=G_{D;\frac{1}{m}}^*$. Without loss of generality, we prove the case with $m=2$ samples. Using a more elaborate but similar argument we can prove it for the general case.

\noindent
Assuming that $D(G(z_1))>D(G(z_2))$, it is easy to show that $w_1>w_2$ and $\log(1-D(G(z_1)))<\log(1-D(G(z_2)))$.
Next, let $n,k$ be positive integers such that $w_1 = \frac{k}{2n} + \varepsilon_1$ and $w_2  = \frac{2n-k}{2n} + \varepsilon_2$, with $\varepsilon_i$ be arbitrarily small constants for $i\in\{1,2\}$. This is possible due to the fact that the set of rational numbers is a dense subset of real numbers. Since $w_1>w_2$ implies $k > n$, then, it holds
\begin{equation*}\small
\begin{aligned}
&w_1 \log(1-D(G(z_1))) + w_2 \log(1-D(G(z_2))) + \varepsilon\\
=&\frac{1}{2n} \big[k\log(1-D(G(z_1))) + (2n-k) \log(1-D(G(z_2)))\big]  + \varepsilon\\
=&\frac{1}{2n} \big[n\log(1-D(G(z_1))) + n\log(1-D(G(z_2)))\\
&+(k-n) (\log(1-D(G(z_1)))-\log(1-D(G(z_2))))\big]  + \varepsilon\\
\leq &\frac{1}{2n} \big[n\log(1-D(G(z_1))) + n\log(1-D(G(z_2)))\big]  + \varepsilon\\
=&\frac{1}{2} \log(1-D(G(z_1))) + \frac{1}{2} \log(1-D(G(z_2))) + \varepsilon,
\end{aligned}
\end{equation*}
for arbitrarily small positive $\varepsilon$. Thus, we prove for $m=2$ that
\begin{equation*} \small
L(G,D;w) \leq L(G,D;\frac{1}{2}). 
\end{equation*}

\noindent
{\bf At equilibrium.} It is straightforward to show that at the Nash equilibrium the weights of WeGAN are uniform. Indeed, it holds that $D(x)=0.5$ for all $x$ and thus
\begin{equation*} \small
w_i = \frac{ \eta^{1-D(G(z_i))}}{\sum_{j=1}^m \eta^{1-D(G(z_j))}} = \frac{\eta^{0.5}}{\sum_{j=1}^m\eta^{0.5}}
= \frac{1}{m} \ .
\end{equation*}
This observation can serve either as a criterion to stop the training process or as an evaluation metric to assess whether or not the training process converged to an optimum. Monitoring the variance of the weights is the simplest statistic for both tasks.

\noindent
{\bf WeGAN generalization.} The proposed algorithm is not exclusive for vanilla GAN and it can be easily extended and applied to any variation of GANs that incorporates a Discriminator mechanism. Therefore, we do not propose just an extension of vanilla GAN but rather a novel training algorithm for general GANs. For instance, we could assign the same formula as in vanilla GAN for the weights for Wasserstein GAN. The presented theoretical analysis still holds for this case.

\vspace{-2mm}
\section{Results}
\vspace{-1mm}
For a fair comparison, we evaluate the performance of the various training algorithms without changing the architecture of the networks. Moreover, with the exception of CIFAR, the presented results are averaged over 1000 iterations.

\subsection{An illustrative example}
We present a benchmark example where the new algorithm converges to the data distribution faster than vanilla GAN. The `real' data are drawn from a mixture of 8 normal distributions with each of the 8 components being equally-probable. The
mean values are equally-distributed on a circle with radius 3 and covariance matrix $I_d$. Moreover, both Generator and Discriminator are fully-connected neural networks with 2 hidden layers and 32 units per layer. The input random variable has a $2$-dimensional standard normal while the output of the Discriminator is the sigmoid function.

\begin{figure}[!htb]
\begin{center}
\includegraphics[width=0.5\textwidth]{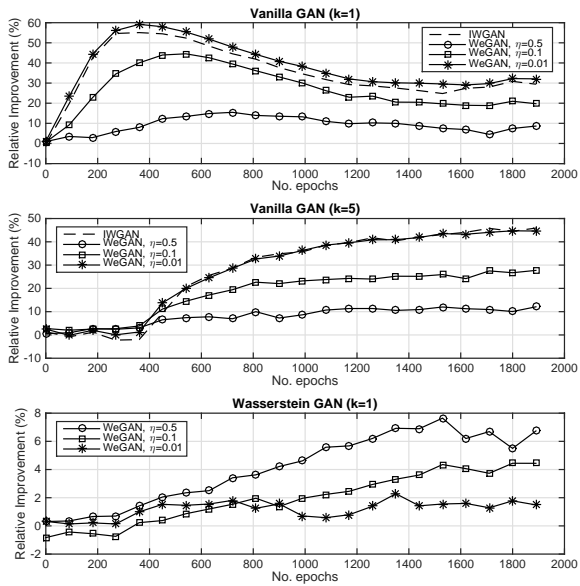}\vspace{-5mm}
\caption{\small Upper \& Middle plot: Relative improvement as a function of the epochs in terms of mean MMD with respect to vanilla GAN for a mixture of 8 Gaussians. Lower values for $\eta$ resulted in improved convergence of WeGAN (lines with circles, squares and stars). Lower plot: Similar to the other plots for Wasserstein GAN. Higher values for $\eta$ gave faster convergence while IWGAN is not applicable.}\vspace{-5mm}
\label{mmd:gmm8:fig}
\end{center}
\end{figure}

The upper and middle plots of Fig. \ref{mmd:gmm8:fig} show the relative improvement of WeGAN with respect of vanilla GAN for various values of $\eta$ (circle, square \& star lines) as a function of the number of epochs. The chosen performance metric is the maximum mean discrepancy (MMD) \cite{gretton2012kernel} which measures the closeness between the real data and the generated ones. The relative improvement is higher at the early stage when only $k=1$ iteration in the training of the Discriminator is performed (upper plot of Fig. \ref{mmd:gmm8:fig}). In contrast, the highest relative improvement occurs closer to the convergence regime when $k=5$ iterations in Discriminator's training are performed (middle plot). For comparison purposes we added IWGAN (dashed line) which also outperforms vanilla GAN but it is slightly worse that WeGAN with $\eta=0.01$. Moreover, there were cases where IWGAN diverges because it produced a weight with infinite value. In the lower plot of Fig. \ref{mmd:gmm8:fig}, we present  the relative performance improvement between the baseline training algorithm for the Wasserstein GAN and the respective weighted variation. We observe that improvements happen but they are less prominent. Additionally, higher values of $\eta$ result in better performance which is the opposite situation when compared with the vanilla GAN.

\subsection{MNIST}
\begin{figure}[!htb]
\begin{center}
\includegraphics[width=0.5\textwidth]{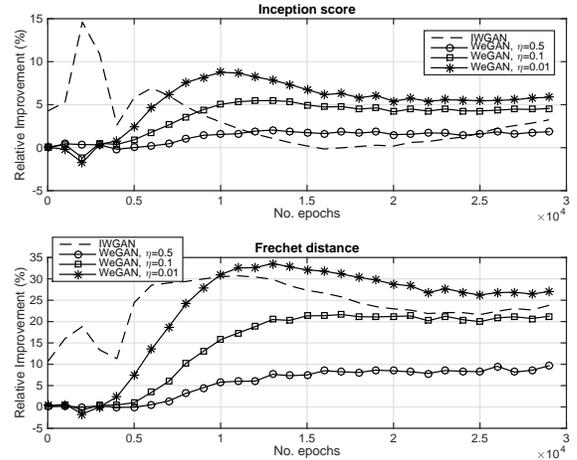}\vspace{-5mm}
\caption{\small Relative improvement as a function of the epochs in terms of IS (upper plot) and FID (lower plot) with respect to vanilla GAN for the MNIST digit dataset. As in the benchmark example, lower values for $\eta$ result in improved convergence of the WeGAN.}\vspace{-5mm}
\label{fid:mnist:fig}
\end{center}
\end{figure}

We extend our experiments on common benchmark MNIST image database of handwritten digits \cite{lecun1998mnist, lecun1998gradient}. In this experiment, a single hidden layer based fully  connected neural network has been used for both Generator and Discriminator with 128 hidden units. Whereas, the input to Generator is set to 100 dimensional standard normal random variables.
Two popular evaluation metrics i.e., Inception Score (IS) \cite{salimans2016improved} and Fr\'echet Inception Distance (FID) \cite{heusel2017gans} are used to quantitatively assess the performance of GANs. Both metrics assume access to a pre-trained classifier and provide an objective score based on the distribution of the sample that is to be evaluated. Overall relative performance, for IWGAN and various versions of WeGAN with respect to vanilla GAN in terms of IS (upper plot) and FID (lower plot) metrics, are presented in Fig. \ref{fid:mnist:fig}. Evidently, WeGAN algorithm outperforms standard vanilla GAN with relative improvement of almost 10\% in IS and 30\% in FID metrics. Results reveal that WeGAN with $\eta=0.01$ has the best improvement when compared to other variations of $\eta$ values which is consistent with the earlier reported results. By examining Fig. \ref{fid:mnist:fig}, we also observe that IWGAN achieves higher relative improvement in the early epochs, however, fails to maintain the performance as oppose to WeGAN at $\eta=0.01$ which procures the best performance. 

\subsection{CIFAR}
\begin{figure}[!htb]
\begin{center}
\vspace{-2mm}
\includegraphics[width=0.5\textwidth]{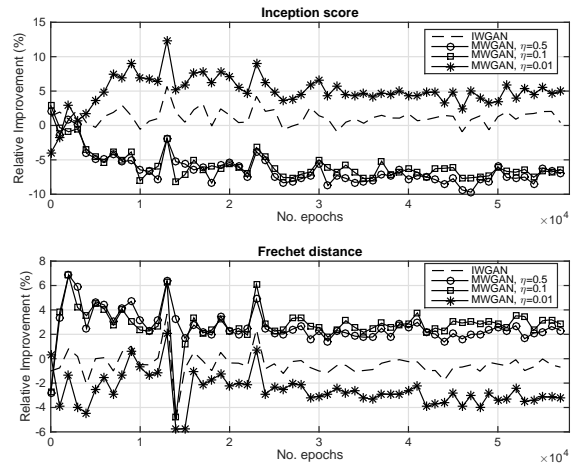}\vspace{-5mm}
\caption{\small Similar to Fig. \ref{fid:mnist:fig} but for the CIFAR-10 dataset. Improvements still happen but they are less prominent while the performance metrics unfortunately produce inconsistent results.}\vspace{-5mm}
\label{cifar10:fig}
\end{center}
\end{figure}

CIFAR-10 is a well studied dataset of natural images \cite{krizhevsky2009learning}. We use this dataset to examine the performance of GANs. For the Generator, we use a deep convolutional network with a single linear layer followed by $3$ convolutional layers. Whereas, the Discriminator has $4$ convolutional layers and $1$ linear layer at the end. Batch normalization is applied to both networks. The input noise with dimensionality of $100$ is drawn from a uniform distribution.
Fig. \ref{cifar10:fig} shows IS (upper plot) and FID (lower plot) scores for the CIFAR-10 dataset in terms of relative improvement with reference to vanilla GAN. It can be observed that the proposed WeGAN with $\eta=0.01$ is preferred over all respective weighted variations in IS score with 5--10\% of improvement. Whereas, WeGAN with $\eta=0.5$ \& $0.1$ both performs comparatively well in FID score. Unfortunately, the performance metrics produce conflicting outcomes making it hard to draw a clear conclusion for this dataset. We also evaluate IWGAN, however, its performance remains approximately the same against the baseline vanilla GAN. 

\section{Conclusions and Future Directions}
Inspired by the \emph{multiplicative weight update method}, we proposed a novel algorithm to train GANs. Results indicated that the performance is improved when compared to the baseline training procedure. Moreover, WeGAN is not restricted to a particular type of GAN but it can be easily applied to any type. As future directions we list a more extensive study in terms of applications and network architectures, a systematic evaluation of the hyper-parameter's behavior as well as extensions towards adding suitable weights to the Discriminator, too.

\section{Acknowledgements}
We would like to thank Yannis Sfakianakis for his help in the implementation of some experiments.

\bibliographystyle{IEEEbib.bst}
\bibliography{bibl}

\end{document}